\documentclass{article}

\usepackage{PRIMEarxiv}

\usepackage[utf8]{inputenc} 
\usepackage[T1]{fontenc}    
\usepackage{hyperref}       
\usepackage{url}            
\usepackage{booktabs}       
\usepackage{amsfonts}       
\usepackage{nicefrac}       
\usepackage{microtype}      
\usepackage{fancyhdr}       
\usepackage{graphicx}       
\graphicspath{{media/}}     

\usepackage{caption}
\usepackage{makecell}
\usepackage[frozencache,cachedir=minted-cache]{minted}
\hypersetup{hidelinks}

\title{WiNGPT-3.0 Technical Report
}

\author{
  WiNGPT Team\thanks{\textbf{Authors are listed in section \ref{sec:authors}}} \\
  \\
  Winning Health AI Research \\
  \texttt{Correspondence to wair@winning.com.cn} \\
}

\begin{document}
\maketitle

\begin{abstract}
Current Large Language Models (LLMs) exhibit significant limitations, notably in structured, interpretable, and verifiable medical reasoning, alongside practical deployment challenges related to computational resources and data privacy. This report focused on the development of WiNGPT-3.0, a 32-billion parameter LLM, engineered with the objective of enhancing its capacity for medical reasoning and exploring its potential for effective integration within healthcare IT infrastructures. The broader aim is to advance towards clinically applicable models. The approach involved a multi-stage training pipeline tailored for general, medical, and clinical reasoning. This pipeline incorporated supervised fine-tuning (SFT) and reinforcement learning (RL), leveraging curated Long Chain-of-Thought (CoT) datasets, auxiliary reward models, and evidence-based diagnostic chain simulation. WiNGPT-3.0 demonstrated strong performance: specific model variants achieved scores of 66.6 on MedCalc and 87.1 on MedQA-USMLE. Furthermore, targeted training improved performance on a clinical reasoning task from a baseline score of 58.1 to 62.5. These findings suggest that reinforcement learning, even when applied with a limited dataset of only a few thousand examples, can improve the accuracy of medical reasoning. Crucially, this demonstration of RL's efficacy with limited data and computation paves the way for more trustworthy and practically deployable LLMs within clinical workflows and health information infrastructures. WiNGPT-3.0 visit information, demos and updates at \href{https://github.com/winninghealth/WiNGPT-3.0}{https://github.com/winninghealth/WiNGPT-3.0}.
\end{abstract}


\section{Introduction}
\label{sec:introduction}

Current Large Language Models (LLMs) demonstrate significant limitations when performing medical reasoning that must be structured, interpretable, and verifiable. While LLMs effectively process large volumes of text, they lack the practical experience and complex diagnostic skills, such as clinical thinking, that human clinicians develop. Furthermore, their reasoning often relies on pattern recognition instead of dynamic, multi-factor clinical assessment; consequently, this reasoning becomes difficult to trace and validate.

Deploying LLMs in real-world healthcare settings adds several hurdles. High computational demands frequently clash with the limited hardware budgets of many institutions, while stringent privacy regulations often require on-premise data processing. Consequently, providers must balance model accuracy, cost, and accessibility to deliver equitable care.

Large reasoning models also tend to hallucinate—producing plausible yet incorrect content—which poses serious clinical risks. Moreover, medical knowledge and guidelines evolve rapidly; without frequent and costly retraining, an LLM’s knowledge base quickly becomes outdated. Adapting LLMs to diverse regional or institutional medical standards presents an additional complex challenge. Such concerns are not merely our observations but a consensus voiced by the healthcare professionals who collaborate with us.

To overcome these obstacles, we introduce WiNGPT, a medical-focused LLM series begun in 2023. The latest version, WiNGPT-3.0, builds on its predecessor WiNGPT-2.8 and employs the 32-billion-parameter Qwen-2.5 architecture \cite{qwen2025qwen25technicalreport}. This scale strikes a balance between robust reasoning power and practical feasibility for on-site deployment. Reinforcement learning on curated clinical case collections encourages the model to emulate clinical thinking. WiNGPT-3.0 is tightly integrated into the WiNEX Health Information System, where it assists clinicians through a triad of safeguards: physician review, structured knowledge bases, and customizable rule templates. These mechanisms improve clinical appropriateness, accuracy, and interpretability, thereby supporting evidence-based practice.

WiNGPT-3.0's development prioritizes deep integration into hospital workflows. Clinicians, administrators, and technical teams achieve this integration through a collaborative process. Together, these teams define requirements, manage system integration, and fine-tune the model using institution-specific data and operational processes. Furthermore, several features ensure its alignment with specific institutional standards and operational processes: pre-configured EMR templates, established rules, and continuous learning from user interactions. This deeply integrated and customizable approach improves the practical utility, reliability, and regulatory compliance of LLMs in clinical settings. A case study illustrating WiNGPT-3.0’s diagnostic-support capabilities—complete with evidence-based search integration—appears in Appendix \ref{app:evidbasedsearch}.

\section{Related Work}

LLMs can significantly enhance their reasoning capabilities by synergistically combining long chain-of-thought (Long-COT) reasoning with reinforcement learning (RL). However, manually annotating the necessary Long-COT data for complex tasks is often prohibitively expensive, posing a critical challenge. To address this data generation bottleneck, researchers are exploring knowledge distillation techniques, applying them to strong foundation models such as DeepSeek-R1 \cite{deepseekai2025deepseekr1incentivizingreasoningcapability}—a leading open-source model known for its reasoning capabilities—to potentially extract high-quality Long-COT datasets. In parallel, to improve training efficiency, algorithms like GRPO \cite{shao2024deepseekmathpushinglimitsmathematical} have substantially reduced the computational complexity and hardware requirements for RL training, making such training more technically feasible in resource-constrained environments. Beyond these approaches focused on data generation and training optimization for specific capabilities, techniques such as model merging constitute an actively evolving area. This methodology, generally implemented during the post-training phase \cite{goddard-etal-2024-arcees}, offers a complementary strategy by combining the parameters of several models, each previously fine-tuned for distinct domain knowledge, to create a more versatile and capable resultant model.

Recent general domain research challenges the widely held assumption of a strong positive correlation between parameter scale and model performance. For instance, 32-billion-parameter models, such as Skywork-OR1-32B-Preview \cite{skywork-or1-2025} and QWQ-32B \cite{qwq32b}, now achieve performance in mathematical reasoning and code generation comparable to significantly larger models like DeepSeek-R1. This success is largely attributed to multi-stage RL training strategies. Such advancements underscore the potential of 'compact models.' They also create opportunities for the substantive integration of medical domain expertise and the adaptation of these models for clinical scenarios. The objective is for such adapted models to maintain high performance while demonstrating enhanced domain-specific comprehension capabilities.

Illustrating this trend in the medical domain, the Huatuogpt-O1 model \cite{chen2024huatuogpto1medicalcomplexreasoning} utilizes RL training augmented by a dynamic scoring mechanism leveraging a GPT-4o validator; this combination has demonstrated strong performance in verifiable medical reasoning tasks. Nevertheless, this approach faces practical challenges, primarily prohibitive validator costs and system stability risks. To address such issues, more recent methods like Search-R1 \cite{jin2025searchr1trainingllmsreason} and RAGEN \cite{wang2025ragenunderstandingselfevolutionllm} integrate knowledge base retrieval mechanisms with RL training paradigms. This integration enables models to autonomously decide when to invoke external knowledge. Such an approach not only substantially enhances model performance but also effectively mitigates the propensity for hallucination common in LLMs. Consequently, these advancements indicate a promising technical pathway for constructing reliable medical AI systems.

\section{Key Features}
\label{sec:key_features}

\begin{itemize}
    \item Achieves state-of-the-art performance in medical-specific reasoning using a 32B parameter model.
    \item Implements a data generation methodology that integrates key techniques such as distillation, reward modelling, and chain-of-thought prompting to efficiently produce high-quality medical datasets under resource constraints.
    \item Introduces a novel reinforcement learning approach tailored for complex clinical reasoning scenarios, including clinical diagnosis simulation and evidence-based search integration.
    \item Develops the WiNGPT-3.0 benchmark for rigorous evaluation of advanced clinical reasoning capabilities.
\end{itemize}



\section{Method}
\label{sec:method}

\begin{figure}
    \centering
    \includegraphics[width=0.65\linewidth]{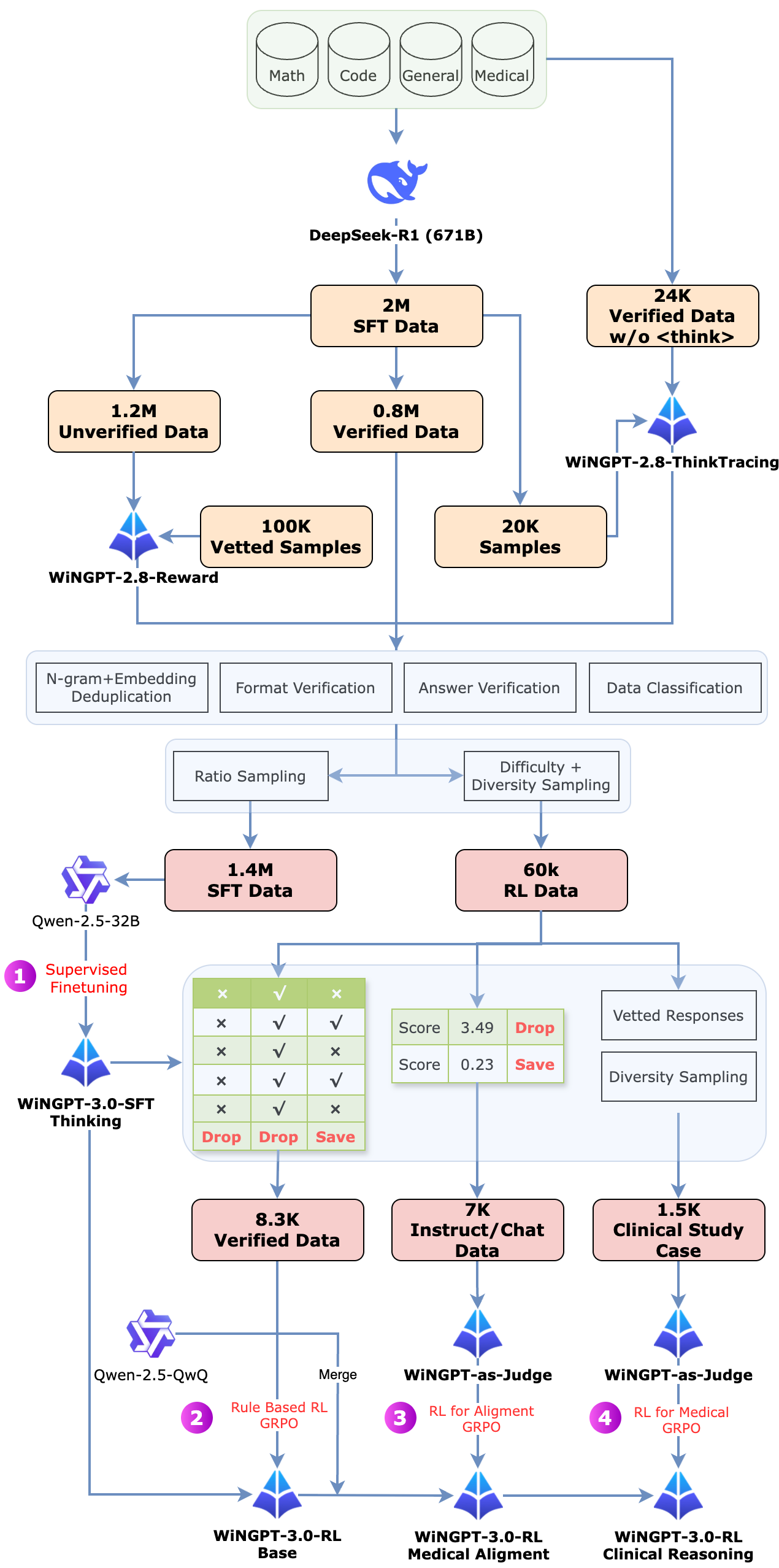}
    \caption{WiNGPT-3.0 Workflow}
    \label{fig:wingpt-3.0-workflow}
\end{figure}

We constructed a large-scale dataset comprising nearly 2 million questions. This dataset encompasses multiple categories (including mathematics, programming, general knowledge, and healthcare) and is provided in both Chinese and English. To generate data suitable for SFT and RL training, we employed a multi-stage data processing pipeline, which included techniques such as long chain-of-thought (Long-COT) answer generation, data filtering and classification, and data sampling. Figure \ref{fig:wingpt-3.0-workflow} illustrates the overall workflow from top to down. Furthermore, to support these data processing efforts, we developed and trained three specialized auxiliary models: a Preference-based Reward Model, a Verifier-based Reward Model, and a Think-Tracing Model. Section \ref{subsec:auxiliary_model} provides further details on these auxiliary models

\subsection{Data Curation}
\textbf{Long Chain-of-Thought Answer Construction}: To acquire long chain-of-thought (Long-COT), we used DeepSeek-R1 \cite{deepseekai2025deepseekr1incentivizingreasoningcapability} for distillation, generating candidate responses for each question. For some medical datasets containing answers but lacking thought processes (approximately 24K), we utilized the Think-Tracing model to generate candidate chains of thought.

\textbf{Data Filtering and Classification}: Considering the verifiability of results, we divided the dataset into two categories: a verifiable dataset (approximately 0.8M) and an unverifiable dataset (approximately 1.2M). In terms of dataset review, we mainly focused on two aspects: firstly, conducting N-gram repetitiveness filtering based on the content of candidate responses, as well as format error filtering based on <think>/</think> tags; secondly, for the verifiable dataset, selecting based on the correctness of the answers; for the unverifiable dataset, using the reward model for answer evaluation and data selection according to the evaluation scores. 

After completing data filtering, we categorized the data based on difficulty into three classes: Basic (low knowledge density or complexity, requiring only basic common sense or simple concepts, direct reasoning process, clear steps); Intermediate (moderate knowledge density or complexity, requiring certain professional knowledge, theories, or formula support, involving multi-step logical deduction); Advanced (high knowledge density or complexity, requiring deep professional knowledge and background theory, involving interdisciplinary integration, complex analysis, or innovative thinking).

\textbf{Data Sampling}: From the Advanced category, we selected 60,000 pieces of data across different domains as the candidate dataset for the reinforcement learning phase. For the SFT training set, we sampled based on difficulty, giving higher sampling ratios to samples of higher difficulty while ensuring diversity balance across all languages and task categories.

\textbf{SFT Data}: After completing sampling, we obtained approximately 1.4 million training data entries in total. Among these, general knowledge data accounted for 50\%, mathematical data 18\%, programming data 14\%, and medical data also accounted for 18\%. This distribution ensures broad domain coverage while also fully considering key professional knowledge.

\textbf{RL Data}: Using the supervised fine-tuned model, we generated N responses for each question in the candidate dataset, removing samples where the model answered either all correctly or all incorrectly. Subsequently, we sampled from lowest to highest number of correct answers among the N responses. The final RL dataset consists of:

\begin{enumerate}
    \item Verifiable dataset of 8.3K entries, primarily including types such as mathematics, medical calculations, and medical quality control;
    \item Unverifiable dataset of 7K entries, mainly comprising types like medical Q\&A, medical diagnosis, medical record generation, medical safety, and general Q\&A;
    \item Clinical thinking dataset of 1.5K entries, aimed at training the model to simulate doctors' clinical diagnostic reasoning processes, thereby assisting clinical decision-making.
\end{enumerate}

\subsection{Training Pipeline}
\begin{enumerate}
    \item[S1] \textbf{Supervised Fine-Tuning}: We performed supervised fine-tuning on the Qwen2.5-32B model \cite{qwen2025qwen25technicalreport} using the standard cross-entropy loss function. This phase involved two epochs on the fine-tuning dataset under a 16K context length configuration.
    \item[S2] \textbf{Reinforcement Learning for General Reasoning}: During the reasoning reinforcement learning stage, the primary goal was to enhance WiNGPT’s logical reasoning capabilities in medical scenarios. We employed a rule-based GRPO \cite{shao2024deepseekmathpushinglimitsmathematical} algorithm to update model parameters using a verifiable dataset. During training, we encouraged the model to prioritize generating shorter responses. To achieve this, we introduced a length penalty mechanism: no penalty was applied if the response was correct and under 8K tokens, while responses exceeding 8K tokens incurred a cosine-based length penalty, with a maximum penalty value of 0.5. After completing this training phase, we explored an extra approach to enhance the model's ability to balance general knowledge and medical knowledge. We merged the WiNGPT-3.0-S2 with QwQ-32B \cite{qwq32b} at a weight ratio of 1:1 before Medical RL was subsequently applied.
    \item[S3] \textbf{Reinforcement Learning for Medical Alignment}: The medical reinforcement learning stage aimed to improve the model’s capability and stability in diverse medical scenarios. To enhance general-domain performance, a small amount of general-domain Q\&A data was incorporated during training. We utilized a model reward mechanism with reference answers, where each question was paired with a reference answer, and Verifier-Based Rewards Model (VRM) was prompted to score the model’s response based on this reference. This approach allowed flexible handling of tasks where rule-based rewards were impractical. At the end, We have completed the training of two models: WiNGPT-3.0-S3 and WiNGPT-3.0-S3-Merged. These models are derived from WiNGPT-3.0-S2 and WiNGPT-3.0-S2-Merged, respectively.
    \item[S4] \textbf{Reinforcement Learning for Clinical Reasoning}: This stage focused on strengthening the model’s clinical decision-making ability by integrating a knowledge base interaction mechanism to improve diagnostic accuracy. Training samples were constructed from real-world clinical diagnostic data, requiring the model to retrieve and reason with knowledge base information during diagnosis. The training adopted the same reward mechanism as the medical reinforcement learning stage, where Verifier-Based Rewards Model (VRM) evaluated the logical completeness of the diagnostic process and the accuracy of conclusions. \cite{qiu2025quantifyingreasoningabilitiesllms}
\end{enumerate}

\subsection{Evidence based Diagnostic Chain}
\begin{figure}
    \centering
    \includegraphics[width=0.9\linewidth]{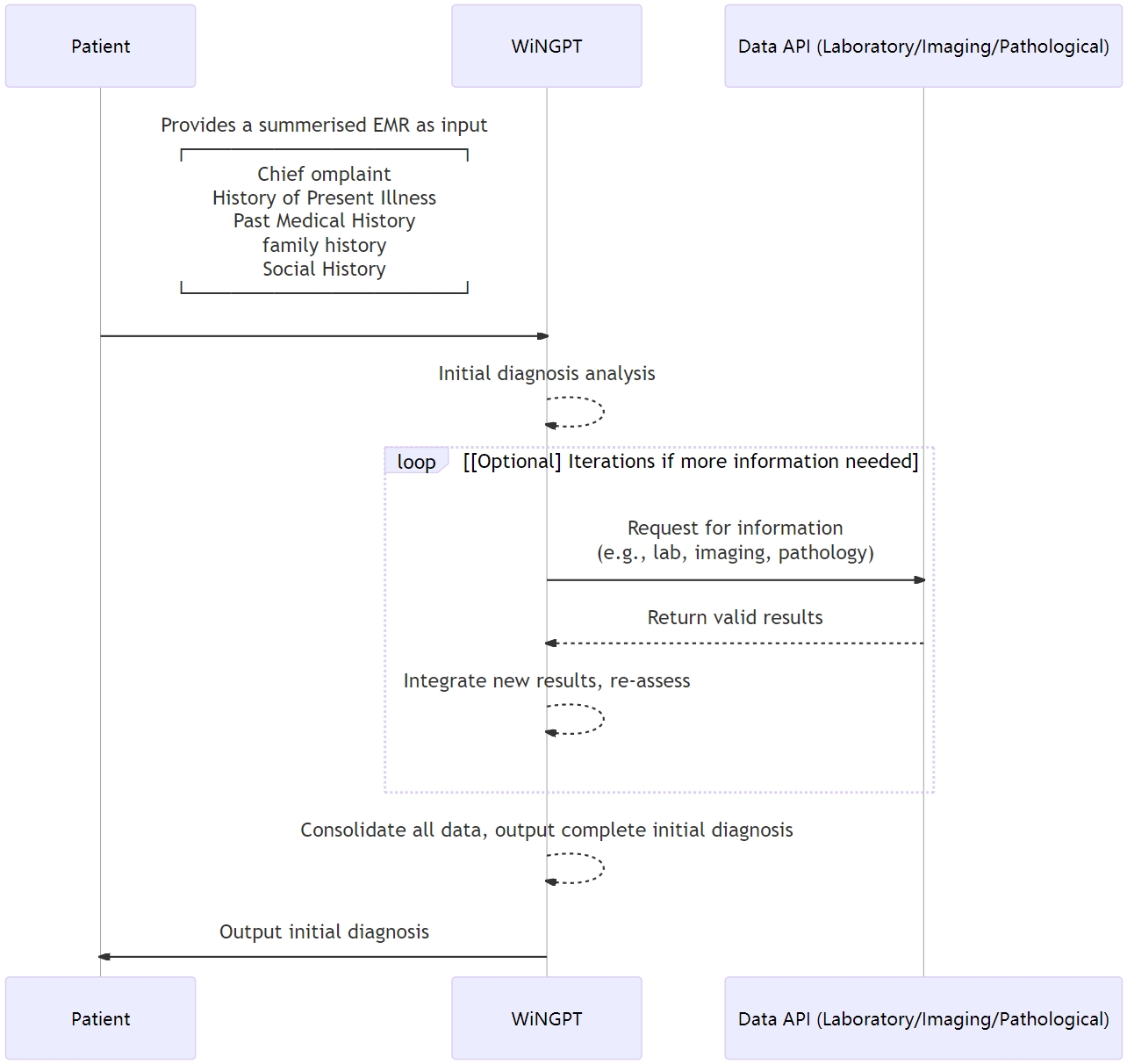}
    \caption{Simulating the reasoning process through a clinical study case}
    \label{fig:simulation_clinical_reasoning}
\end{figure}

In medical education and training, evidence-based thinking plays a crucial role. Diagnosis and treatment decisions can only be applied to patients after identifying supporting evidence and undergoing thorough discussion and evaluation. In reinforcement learning, this approach can be analogized to mathematical problems, where the model is required to provide the final answer in a specified format (e.g., within a designated box), enabling reliable rule-based verification of correctness. Based on this concept, we have simulated a clinical diagnostic workflow as shown in Figure \ref{fig:simulation_clinical_reasoning} and constructed an evidence-based auxiliary diagnosis dataset for training clinical reasoning models.

For this purpose, we selected comprehensive data from an electronic medical record (EMR) database, including key components of patient records: chief complaint, present illness, medical history, physical examination, relevant auxiliary tests, and final diagnostic results. Inspired by the work of \cite{qiu2025quantifyingreasoningabilitiesllms}, we designed a multi-step auxiliary diagnostic pathway consisting of three primary stages: initial consultation summary, information requirements and diagnosis.

We first generated the initial consultation summary, incorporating the patient's chief complaint, present illness, and medical history. Additional data, such as physical examination and auxiliary tests, were organized into a separate examination module. The reasoning model evaluates whether an auxiliary diagnostic conclusion can be derived based on the initial consultation summary. If an auxiliary diagnosis cannot be determined, additional relevant examinations are required.

To facilitate this process, we designed an autonomous agent that queries the necessary data and provides the corresponding examination results. This iterative process repeats multiple times until a final auxiliary diagnosis is successfully generated.

\subsection{Auxiliary Model}
\label{subsec:auxiliary_model}

\textbf{Preference-Based Rewards}:
Our reward model, WiNGPT-32B-RM, is initialized from the WiNGPT-2.8-32B model (itself based on Qwen-2.5-32B \cite{qwen2025qwen25technicalreport}). We modified its final output layer to a linear head that directly outputs a single scalar logit. This scalar logit quantifies the model's preference strength for a given "prompt + response" pair. For training WiNGPT-32B-RM, we employ the Scaled Bradley-Terry (BT) Loss \cite{lambert-etal-2025-rewardbench}, optimizing the model using human-annotated pairwise comparison data. The introduction of a scaling factor in this loss function enhances the model's ability to distinguish between subtle variations in human preference, improves training stability, and leads to more accurately calibrated reward signals. Our internal experiments indicated that this approach yields slightly better performance compared to the standard BT loss \cite{Hunter2003MMAF}. To train WiNGPT-32B-RM, we assembled a comprehensive training corpus by combining internally curated data with existing general-purpose and domain-specific preference datasets. This strategy aims to ensure both strong medical alignment and broad general reward modelling capability. Our internal contribution consists of 100K meticulously selected, high-quality medical preference pairs. For each prompt in this internal dataset, a preferred response was selected by annotators from multiple model generations, and a corresponding less-preferred (negative) response was sampled from generations of an open-source model. In addition to this internal set, the training data for WiNGPT-32B-RM incorporates:

\begin{itemize}
    \item General-purpose preference datasets: HelpSteer3 \cite{wang2025dedicatedfeedbackeditmodels} and INF-ORM-Preference-Magnitude-80K \cite{INF-ORM-Llama3.1-70B}.
    \item Medical-specific preference annotations from publicly available evaluation datasets: MedAlign-ModelEval (36k model-generated preference annotations) and MedAlign-HumanEval (11k human-generated preference annotations), which provide valuable domain-specific training signals.
\end{itemize}

Finally, We evaluated WiNGPT-32B-RM on two distinct benchmarks: MedRewardBench, an internally developed benchmark specifically targeting medical preference alignment, and the publicly available RewardBench, which measures general reward model performance. WiNGPT-32B-RM achieved a score of 96.4 on MedRewardBench and 92.4 on RewardBench, demonstrating strong performance in both domain-specific and general contexts.

\textbf{Think-Tracing}: To efficiently leverage verifiable data that lacks a chain of thought, we extracted healthcare data containing high-quality Long-COT annotations, using questions and answers as inputs and the chain-of-thought as outputs, forming a 20K-sample training set for the thought trajectory model based on WiNGPT-2.8. This method is particularly valuable for creating training data for out-of-domain, verifiable data, and for converting existing non-reasoning datasets or model outputs into rich, traceable reasoning datasets, thus greatly enhancing scalability in data creation. Most of the data we've verified manually needs further validation through experiments to confirm its effectiveness.

\textbf{Verifier-Based Rewards}: We conducted high-quality answer sampling based on an unverifiable dataset, collecting 90K single-turn dialogues and 7K multi-turn dialogues. The processing workflow consists of the following steps: First, we utilized answers generated by DeepSeek-R1 \cite{deepseekai2025deepseekr1incentivizingreasoningcapability} as reference answers, removing all CoT content. Concurrently, we adopted a random selection strategy to generate evaluation answers using both fine-tuned model and model from the reasoning reinforcement learning phase, with CoT content similarly removed. We then constructed a question set through carefully designed prompts and employed DeepSeek-R1 \cite{deepseekai2025deepseekr1incentivizingreasoningcapability} to perform differentiated evaluations between reference answers and evaluation answers, implementing a binary scoring mechanism (0/1) for quality assessment. The high-quality outputs devoid of CoT content were subsequently used as training data. Through two training epochs of fine-tuning, we successfully established a Verifier-Based Rewards Model (VRM). Throughout this process, we maintained concise answer outputs, with all CoT content systematically removed during the preprocessing phase.

\subsection{Training Configuration}
During the supervised fine-tuning (SFT) phase, we employed the ZeRO-3 \cite{10.5555/3433701.3433727} memory optimization strategy within the DeepSpeed \cite{10.1145/3394486.3406703} framework. For optimization in this phase, the AdamW optimizer was configured with an initial learning rate of $1 \times 10^{-5}$. This was combined with a linear warmup period over 500 training steps, followed by a cosine annealing schedule that decayed the learning rate to 10\% of its initial value. The model underwent SFT for a total of 2 epochs. Subsequently, for the RL phase, we utilized the GRPO \cite{shao2024deepseekmathpushinglimitsmathematical} algorithm, implemented via the Verl \cite{Sheng_2025} framework. The RL training employed a per-step batch size of 128 samples, with 12 rollouts generated per prompt.

\section{Evaluation}

\subsection{WiNGPT-3.0 Benchmark}
WiNGPT-3.0 Benchmark is a multi-task evaluation benchmark designed to assess the comprehensive capabilities of LLMs in medical and high-level cognitive tasks. The benchmark framework aims to comprehensively reflect the reasoning, computation, and language understanding abilities of models in real-world clinical applications, covering multiple dimensions from professional medical knowledge application to complex mathematical problem-solving.

The design of WiNGPT-3.0 Benchmark follows these principles:
\begin{itemize}
    \item \textbf{Clinical Relevance}: Ensures that the selected tasks are highly aligned with real-world medical decision-making, covering key aspects of daily clinical workflows.
    \item \textbf{Professionalism and Challenge}: Incorporates tasks with a high degree of specialization, requiring models to possess professional clinical knowledge and reasoning capabilities.
    \item \textbf{Cross-task Generalization Assessment}: Tests models' generalization and task-transfer abilities by combining various types of tasks.
\end{itemize}

The WiNGPT-3.0 benchmark consists of a diverse set of task categories as following.

\begin{itemize}
    \item MedCalc \cite{khandekar2024medcalcbenchevaluatinglargelanguage} includes 2,811 questions, serving as a dataset for evaluating model performance on clinical calculation tasks, covering 55 different types of medical calculations.
    \item MedReMCQ contains 700 medical reasoning multiple-choice questions involving diagnostic tests, medical record quality control, and preliminary diagnosis. Examples can be found in Appendix \ref{app:medremcq_examples}.
    \item CMMLU \cite{li-etal-2024-cmmlu} provides 11,582 questions and is a benchmark for evaluating Chinese language understanding across multiple tasks.
    \item MATH-500 \cite{lightman2023letsverifystepstep} consists of 500 advanced mathematical problems primarily used to test and train mathematical reasoning abilities.
    \item MedQA-USMLE \cite{app11146421} offers 1,273 questions, forming a large-scale open domain question answering dataset from medical exams.
    \item MedMCQA \cite{pmlr-v174-pal22a} features 4,183 questions and is a large-scale multi-subject multi-choice dataset for medical domain question answering.
    \item PubMedQA \cite{jin2019pubmedqadatasetbiomedicalresearch} comprises 1,000 questions, designed as a dataset for biomedical research question answering.
\end{itemize}

\subsection{Evaluation Metrics}
We employ a set of specific evaluation metrics, each chosen to best reflect the original benchmark's intent.

\begin{itemize}
    \item MedCalc \cite{khandekar2024medcalcbenchevaluatinglargelanguage} uses accuracy as the metric which is based on whether the model’s computed answer falls within the upper and lower bounds of the standard answer.
    \item MedReMCQ is a dataset we developed in-house, tailored to medical reasoning scenarios. It is presented in a multiple-choice format and assessed using the Micro-F1 score.
    \item CMMLU \cite{li-etal-2024-cmmlu}), MATH-500 \cite{lightman2023letsverifystepstep}, MedQA-USMLE \cite{app11146421}, MedMCQA \cite{pmlr-v174-pal22a} and PubMedQA \cite{jin2019pubmedqadatasetbiomedicalresearch} use accuracy as the evaluation metric.
\end{itemize}

\subsection{Result}

We evaluated these strategies on two general reasoning/math benchmarks (CMMLU and MATH-500) and five medical benchmarks (MedCalc, MedReMCQ, MedQA-USMLE, MedMCQA, and PubMedQA), comparing them with several SOTA 32B-RL models. The evaluation results are summarized in Table \ref{tab:model_performance}.

\begin{table}[htbp]
    \renewcommand{\arraystretch}{1.5}
    \centering
    \setlength{\tabcolsep}{4pt}
    \begin{tabular}{@{}l*{7}{c}@{}}
        \toprule
        \thead{\textbf{Model Name}} &
        \thead{\rotatebox[origin=l]{45}{\textbf{CMMLU}}} &
        \thead{\rotatebox[origin=l]{45}{\textbf{MATH-500}}} &
        \thead{\rotatebox[origin=l]{45}{\textbf{MedCalc}}} &
        \thead{\rotatebox[origin=l]{45}{\textbf{MedReMCQ}}} &
        \thead{\rotatebox[origin=l]{45}{\textbf{MedQA-USMLE}}} &
        \thead{\rotatebox[origin=l]{45}{\textbf{MedMCQA}}} &
        \thead{\rotatebox[origin=l]{45}{\textbf{PubMedQA}}} \\
        \midrule
        DeepSeek-R1-Distill-Qwen-32B & 83.0 & 96.3 & 57.7 & 65.3 & 81.9 & 65.5 & 76.1 \\
        OpenThinker2-32B             & 86.2 & 96.6 & 58.5 & 68.8 & 85.1 & 68.7 & 78.0 \\
        Light-R1-32B-DS              & 86.2 & 97.2 & 48.7 & 70.3 & 83.7 & 67.9 & 76.2 \\
        QwQ-32B                      & \textbf{87.4} & \textbf{97.5} & 51.4 & 73.5 & 85.8 & 71.2 & 77.9 \\
        WiNGPT-3.0-S3             & 85.1 & 94.0 & \textbf{66.6} & \textbf{74.7} & 85.7 & 69.9 & \textbf{78.2} \\
        WiNGPT-3.0-S3-Merged      & 86.1 & 95.6 & 66.2 & 74.0 & \textbf{87.1} & \textbf{71.6} & 78.0 \\
        \bottomrule
    \end{tabular}
    \caption{Performance comparison of different models across various benchmarks.}
    \label{tab:model_performance}
\end{table}

Both versions of WiNGPT-3.0 demonstrate competitive performance in general reasoning and mathematics, with the S3-Merged model achieving a notable boost in MATH-500 and competitive CMMLU scores, suggesting the merged model retains strong foundational capabilities.

In the medical domain, both strategies yield significant improvements over baseline models. WiNGPT-3.0-S3-Merged shows particularly strong performance in MedQA-USMLE and MedMCQA, indicating that merging with QwQ-32B followed by medical fine-tuning can effectively enhance domain-specific knowledge transfer.

To further enhance the model's performance in clinical reasoning scenarios, RL training was applied in S4 using a 1.5K clinical thinking dataset. This dataset simulates the real-world medical consultation process, encompassing key steps such as model-driven reasoning, initial consultation summary, information requirements, and diagnosis. Through this pipeline, the model gains improved reasoning capabilities in simulated clinical settings. Experimental results demonstrate that this application of RL in Step 4 leads to a notable improvement in clinical reasoning tasks, with the WiNGPT-3.0-S4 model achieving a score of 62.5, compared to 58.1 for the WiNGPT-3.0-S3 model.

\section{Discussion}

RL markedly improved alignment with clinical-reasoning principles. WiNGPT-3.0-S3 outperformed the base QwQ-32B on MedCalc (66.6 vs. 51.4) and MedReMCQ (74.7 vs. 73.5), indicating stronger medical reasoning ability. Merging with a strong model (QwQ-32B) prior to RL fine-tuning (WiNGPT-3.0-S3-Merged) yielded substantial benefits; its superior performance across diverse benchmarks—medical (MedQA-USMLE: 87.1; MedMCQA: 71.6), general (CMMLU: 86.1), and mathematical reasoning (MATH-500: 95.6)—indicates that this integration effectively preserves general capabilities while facilitating domain-specific adaptation. Nonetheless, current benchmarks still under-represent real-world complexity: they emphasise pattern matching, overlook temporal factors, and offer limited probes for step-wise reasoning.

Beyond numerical scores, WiNGPT-3.0 may influence patient–clinician interactions in several ways:

\begin{itemize}
    \item \textbf{Patient perspective}: When used as an \textit{educational companion}—rather than an independent diagnostician—the model can give lay users a plain-language preview of plausible conditions, supporting evidence, and follow-up questions. This early framing may reduce anxiety, encourage timely medical visits, and help patients describe symptoms more precisely, making subsequent consultations more focused.
    \item \textbf{Clinician perspective}: Because patients arrive better informed, physicians must be ready to validate, refine, or correct AI-generated hypotheses. Competent engagement with these “AI-augmented” patients is essential to maintain trust and professional credibility.
\end{itemize}

\textbf{Limitations and safety}: Bias in training data and hallucinations remain open challenges. Evidence-retrieval pipelines reduce—but cannot eliminate—errors because search results themselves may be outdated or incomplete. We therefore require licensed clinicians to review all outputs and discourage unsupervised use by non-professionals.

Again, applying LLMs in healthcare system incorrectly could cause serious harm \cite{TayyarMadabushi73}; therefore, these models must be used carefully. We believe that open sharing of this report will help share knowledge and improve AI research for healthcare.

\section{Conclusion}
This technical report details WiNGPT-3.0, outlining the methodologies for general and medical domain data curation, alongside strategies for RL data generation and its associated training paradigms. The report further elaborates on the fine-tuning and RL techniques employed, leveraging the specifically constructed datasets detailed herein. Benchmark evaluations demonstrate that WiNGPT-3.0 not only surpasses its predecessor but also exhibits competitive, and in several key tasks, superior performance when compared to general-purpose models of an equivalent parameter scale. A significant contribution of this work is the enhanced focus on medical domain data curation and a preliminary investigation into the novel application of reinforcement learning to imitate clinical thinking. This provides a valuable foundation and resource for researchers aiming to advance AI-driven applications in healthcare.

Future research will concentrate on developing advanced training methodologies for complex medical scenarios, improving model reliability, safety, and interpretability, exploring multi-modal data integration, and rigorously evaluating its potential to deliver positive societal impact, particularly within the healthcare sector. We aim to further reduce the cost of running LLMs, enabling their wider application in real-world scenarios.

\section{Authors}
\label{sec:authors}
Boqin Zhuang, Chenxiao Song, Huitong Lu, Jiacheng Qiao, Mingqian Liu, Mingxing Yu, Ping Hong, Rui Li, Xiaoxia Song, Xiangjun Xu, Xu Chen, Yaoyao Ma, Yujie Gao

\bibliographystyle{unsrt}
\bibliography{main}

\newpage

\appendix
\section*{Appendix}

\section{MedReMCQ Examples}
\label{app:medremcq_examples}

\begin{figure}[H]
    \centering
    \includegraphics[width=0.6\linewidth]{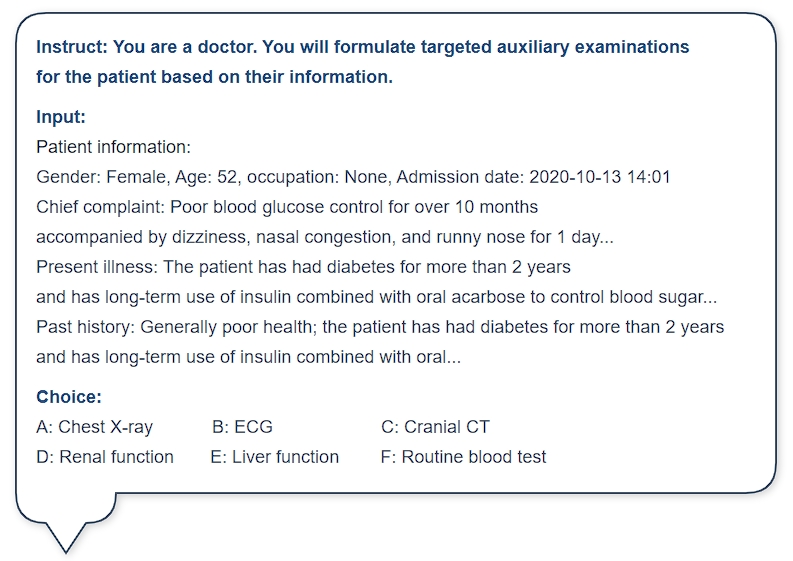}
    \caption{Auxiliary Examinations Selection}
    \label{fig:example1}
\end{figure}

As shown in Figure \ref{fig:example1}, the patient is a 52-year-old female presenting with poor blood glucose control for over 10 months, accompanied by dizziness, nasal congestion, and runny nose for one day. She has a history of diabetes for over two years and has been using insulin combined with acarbose for glycemic control. Based on her condition, the physician needs to conduct targeted auxiliary examinations.

\begin{figure}[H]
    \centering
    \includegraphics[width=0.6\linewidth]{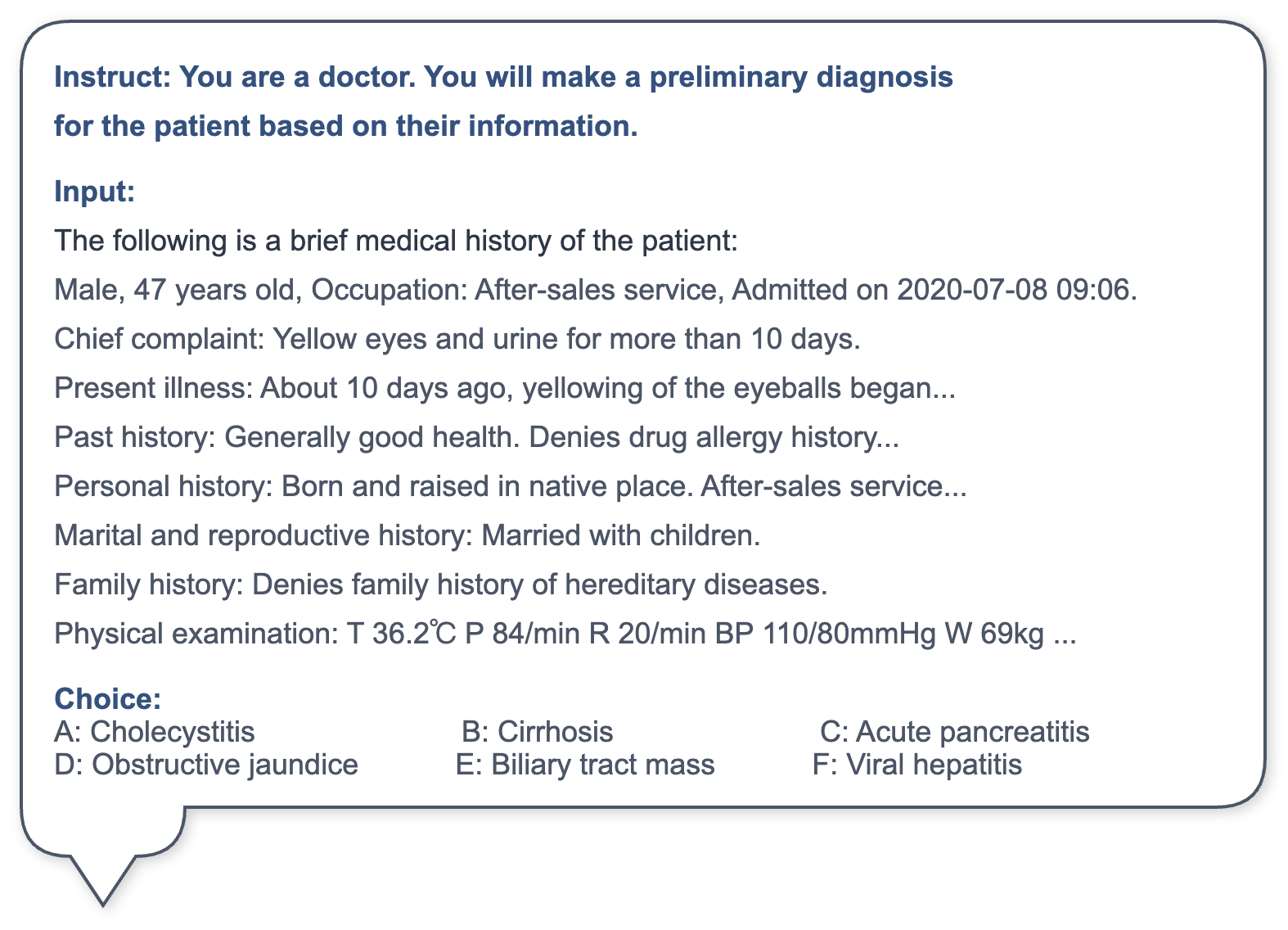}
    \caption{Preliminary Diagnosis}
    \label{fig:example2}
\end{figure}

As shown in Figure \ref{fig:example2}, the patient is a 47-year-old male who complains of yellowing of the eyes and urine for more than 10 days, with a relatively acute onset. His body temperature is normal on physical examination, and there are no obvious signs of abdominal pain or fever. He reports generally good health and no family history of hereditary diseases. The physician needs to make a preliminary diagnosis.

\section{The Prompt for Verifier-Based Rewards}

\begin{figure}[H]
    \begin{minted}[frame=single, breaklines]{markdown}
    You are a professional evaluation expert who must assess the quality of the "predicted answer" based on the following four core elements:
    
    - **Dialogue History** (contextual information)
    - **Current Question** (the specific request made by the user)
    - **Excellent Answer** (a high-quality reference answer that has been reviewed)
    - **Predicted Answer** (the answer to be evaluated)
    
    ### Scoring Criteria:
    - **1 point**: Indicates high-quality answers that are equivalent or close to the "excellent answer," meeting user needs;
    - **0 points**: Indicates low-quality answers that contain hallucinations, omissions, errors, or fail to meet user needs;
    
    > Note: The "excellent answer" has undergone rigorous review and is considered to be of high standard quality, serving as a benchmark for judgment.
    
    Please output your evaluation results in the following structure:
    
    ---
    
    ### Dialogue History (in chronological order, from earliest to latest)
    ```
    {Insert dialogue history}
    ```
    
    ### Current Question
    ```
    user: {Insert original question}
    ```
    
    ### Excellent Answer (Reference Answer)
    ```
    assistant: {Insert excellent answer}
    ```
    
    ### Predicted Answer (Answer to Evaluate)
    ```
    assistant: {Insert predicted answer}
    ```
    
    ---
    
    ### Evaluation Analysis
    
    [Perform item-by-item comparative analysis here]
    
    ---
    
    ### Predicted Answer Evaluation Score
    
    \\boxed{Prediction answer evaluation score}
    \end{minted}
\end{figure}

\section{Evidence-Based Search Integration: A Case Study}
\label{app:evidbasedsearch}

Due to the rapid expansion of medical knowledge, traditional search tools are often inefficient in locating critical evidence within extensive medical literature and clinical data, leading to information overload that can compromise the efficiency and accuracy of clinical decision-making. Evidence-Based Search Integration is presented as a promising approach to mitigate these challenges. Its core value lies in systematically integrating authoritative medical evidence with real-time patient data to provide clinicians with precise, interpretable, and research-backed decision support. This system is designed with the aim to help optimize treatment strategies, contribute to reducing misdiagnosis rates, and promote more rational resource allocation. The technical implementation strategy detailed herein comprises three primary stages:

\begin{itemize}
    \item \textbf{Development of an AI-powered intelligent search engine}: This involves creating a search engine that integrates global authoritative medical databases (e.g., PubMed, Cochrane Library, UpToDate) with localized medical data resources (e.g., electronic medical records, regional health archives, structured knowledge graphs). Seamless integration is facilitated by standardized interfaces and data protocols. Advanced AI algorithms, such as Embedding and Reranker models, are employed for semantic understanding and relevance ranking of search results, prioritizing high-quality, peer-reviewed medical guidelines and research literature.
    \item \textbf{Task planning and complex query parsing using LLMs}: To address complex or ambiguous clinical queries, the system incorporates LLMs for task decomposition and logical reasoning. These models identify key terms and contextual relationships within queries, breaking them down into executable subtasks. Natural Language Processing (NLP) techniques and medical knowledge graphs are utilized to translate user input into precise search commands, facilitating rapid matching with relevant literature, guidelines, and clinical cases. The system also supports multilingual interaction.
    \item \textbf{Knowledge fusion and traceable answer generation}: In this stage, the system combines contextual knowledge bases with LLM capabilities to produce structured, explainable medical responses. Each response includes citations or references to source materials to ensure traceability and support verifiability. Personalized and dynamically updated recommendations are generated based on the latest clinical evidence and individual patient data (e.g., laboratory tests, imaging reports). Machine learning algorithms are employed to continuously refine and optimize information ranking logic. A comprehensive list of references, citing applicable medical guidelines, databases, or journal sources, is provided with each response to enhance credibility and transparency.
\end{itemize}

In summary, the case study describes a three-step, integrated process—data integration, intelligent parsing, and traceable answer generation—designed to efficiently transform raw information into actionable clinical insights. This system aims to enhance physicians' information processing efficiency and support diagnostic accuracy, thereby contributing to more evidence-informed, personalized patient care and laying a technological foundation for advancing smart healthcare.

\end{document}